\documentclass[10pt,twocolumn,letterpaper]{article}

\usepackage[pagenumbers]{cvpr} 

\usepackage{amsmath}
\usepackage{algpseudocode}
\usepackage{array}
\usepackage{multirow}
\usepackage{dsfont}
\usepackage[font={small}]{caption}
\usepackage{wrapfig,lipsum,booktabs}
\usepackage{adjustbox}
\usepackage[ruled, lined, linesnumbered, commentsnumbered, longend]{algorithm2e}
\usepackage{listings}

\usepackage{makecell}
\usepackage{pifont}

\definecolor{cvprblue}{rgb}{0.21,0.49,0.74}
\usepackage[pagebackref,breaklinks,colorlinks,allcolors=cvprblue]{hyperref}

\title{EditAR: Unified Conditional Generation with Autoregressive Models}

\author{Jiteng Mu\textsuperscript{1}, \quad
Nuno Vasconcelos\textsuperscript{1}, \quad
Xiaolong Wang\textsuperscript{1,2} \\
\textsuperscript{1}UC San Diego, \textsuperscript{2}NVIDIA
}

\begin{document}

\twocolumn[{
\vspace{-1em}
\maketitle
\vspace{-1em}

\begin{center}
    \centering
    \vspace{-0.20in}
    \includegraphics[width=\linewidth]{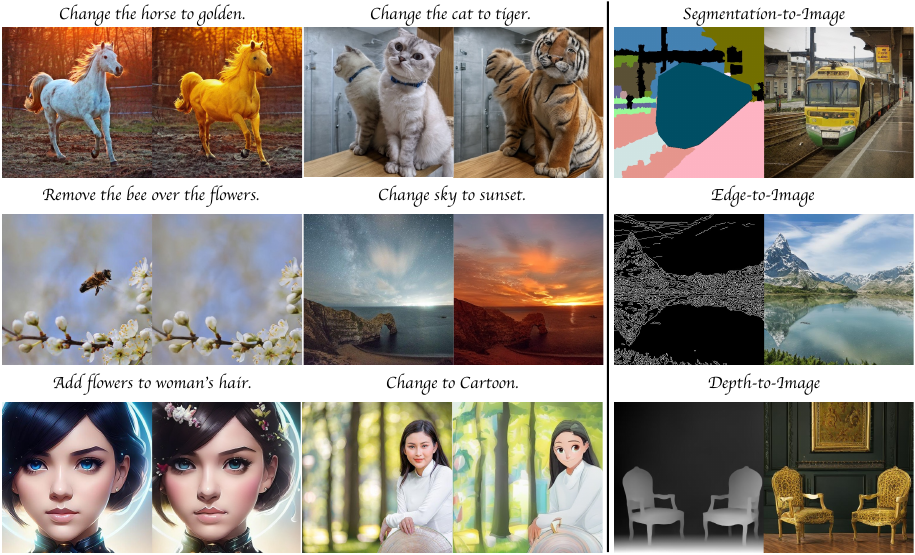}
    \vspace{-0.20in}
    \captionof{figure}{
    We propose EditAR, a unified conditional autoregressive model for diverse conditional generation tasks. We demonstrate that without task-specific designs, a single autoregressive model achieves strong performance across diverse tasks, including texture manipulation, object replacement, object removal, local editing, canny-to-image, depth-to-image, and segmentation-to-image.
    }
    \label{fig:teaser}
\end{center}
}]

\begin{abstract}
Recent progress in controllable image generation and editing is largely driven by diffusion-based methods. Although diffusion models perform exceptionally well in specific tasks with tailored designs, establishing a unified model is still challenging. In contrast, autoregressive models inherently feature a unified tokenized representation, which simplifies the creation of a single foundational model for various tasks. In this work, we propose \textbf{EditAR}, a single unified autoregressive framework for a variety of conditional image generation tasks, e.g., image editing, depth-to-image, edge-to-image, segmentation-to-image. The model takes both images and instructions as inputs, and predicts the edited images tokens in a vanilla next-token paradigm. To enhance the text-to-image alignment, we further propose to distill the knowledge from foundation models into the autoregressive modeling process. We evaluate its effectiveness across diverse tasks on established benchmarks, showing competitive performance to various state-of-the-art task-specific methods. Project page: \url{https://jitengmu.github.io/EditAR/}

\end{abstract}

\section{Introduction}
\label{sec:introduction}

Recent advances in conditional image generation are predominantly driven by high capacity text-to-image diffusion models~\cite{ramesh2021dalle,saharia2022imagen,rombach2021highresolution,ho2020ddpm,song2020ddim,nichol2021glide,ho2022cascadediffusion,peebles2023dit}. These models serve as a strong image prior, that can be specialized to individual tasks, facilitating remarkable progress in applications like inpainting~\cite{Lugmayr2022repaint,yang2023paint,wasserman2024paint}, personalization~\cite{gal2022textual,kumari2023multi,ruiz2023dreambooth}, image editing~\cite{hertz2022prompt,brooks2022instructpix2pix,mokady2023null,ju2023pnpinversion,mu2024eie}, and image translation ~\cite{zhang2023adding,li2024controlnetplusplus}. While these approaches excel at each individual task, the resulting variations in architectural designs and learning objectives make it challenging to integrate multiple tasks within a single framework. Ideally, a single conditional model would excel across all tasks, simplifying both implementation and deployment.

Large-scale autoregressive models have recently gained attention as a promising path towards this goal~\cite{yu2022vector,yu2022parti,sun2024llamagen,keyu2024var}. Unlike diffusion models, they naturally provide a unified, token-based, framework to blend diverse inputs. This unified representation offers substantial promise for creating a single model capable of performing a wide range of image synthesis and manipulation tasks within a unified architecture. While recent works~\cite{sun2024llamagen,keyu2024var} demonstrate remarkable text-to-image generation performance, it is still unclear whether such models are well-suited as foundational architectures for broader conditional generation tasks. While a few autoregressive variants~\cite{chang2022maskgit} have demonstrated the possibility of image inpainting by modifying the next-token prediction paradigm, the efficacy of next-token prediction has not yet been demonstrated for image editing.

In this work, we investigate the feasibility of building a single autoregressive model that unifies various conditional image generation tasks. We introduce \textbf{EditAR}, a novel autoregressive model that, for the first time, integrates various image manipulation and image translation tasks. Like prior text-to-image autoregressive models~\cite{sun2024llamagen,keyu2024var,chang2022maskgit}, it consists of two stages: a VQVAE~\cite{Oord2017vqvae,razavi2019vqvae2}, that maps image patches into tokens indices, and an autoregressive transformer~\cite{esser2021taming} that models the categorical distribution of output tokens given both texts and images as inputs. As illustrated in Figure~\ref{fig:teaser}, we show that that this achieves promising performance {\it across a significant variety of tasks\/}. 

EditAR builds primarily on Llamagen~\cite{sun2024llamagen}, a text-to-image autoregressive model based on the Llama2~\cite{hugo2023llama,hugo2023llama2} architecture that has demonstrated impressive image generation capabilities. However, due to the lack of a conditional image input, Llamagen does not support tasks like image manipulation or translation. To allow this, we adapt the architecture by prefilling the model with image tokens from a conditioning input image, along with additional positional embeddings. This is complemented by an auxiliary distillation loss, based on DINOv2~\cite{caron2021dino,oquab2024dinov2}, which reinforces the visual coherence of the images synthesized by the autoregressive model. Empirical results show that injecting this visual prior enhances alignment between the generated images and the input text. 

At inference, given an image and corresponding editing instructions, output tokens are generated by the standard next-token prediction paradigm. To enhance image quality and text-image alignment, classifier-free guidance~\cite{ho2022cfg} is applied to both image and texts. To our knowledge, this paper provides the first evidence for the feasibility of using next-token prediction autoregressive models for conditional generation on large benchmarks. Furthermore, by solving diverse conditional image generation tasks, it paves the way for a new class of approaches to unified conditional generation. Our contributions are as follows:

\begin{itemize}
    \item We introduce a new autoregressive framework EditAR, that is jointly trained on various image manipulation and image translation tasks, and demonstrates promising potential towards building a unified conditional image generation model.
    \item A distillation loss is introduced to enhance the semantics in the learning of autoregressive models.
    \item Experiments show that the proposed method demonstrates strong performance on a variety of tasks, including texture manipulation, object replacement, object removal, local editing, canny-to-image, depth-to-image, and segmentation-to-image.
\end{itemize}

\section{Related Work}
\label{sec:related work}

\subsection{Controllable Image Diffusion} 
Text-to-image diffusion models~\cite{ho2020ddpm,james2023dalle3,patrick2024sd3,chen2023pixart,podell2023sdxl,nichol2021glide,saharia2022imagen,peebles2023dit,ramesh2022dalle2,ho2022cascadediffusion,ho2022cfg,lu2022dpmsolver,dhariwal2021adm,song2019scoredistillation,song2020ddim,song2023consistency,rombach2021highresolution} have revolutionized image generation, significantly improving the realism of the generated visuals. Nevertheless, adapting these models for tasks like image editing poses notable challenges, as synthesized images can include artifacts, fail to follow the editing instruction, or fail to remain faithful to the conditioning image.

One popular approach~\cite{meng2021sdedit,wallace2023edict,hertz2022prompt,mokady2023null,ju2023pnpinversion,parmar2023p2pzero,daiki2023negativepromptinversion} is to frame editing as a two stage process: the conditioning image is first inverted into noise, which is then run through the forward diffusion chain conditioned by the editing instructions. 
For example, SDEdit~\cite{meng2021sdedit} adds noise to the image, which is then progressively denoised via a stochastic differential equation (SDE). This, however, can struggle to reconstruct the image details that are supposed to remain unaltered. More advanced techniques improve both the inversion and content-preserving generation quality. To improve the reconstruction quality, optimization-based inversion methods~\cite{mokady2023null,ju2023pnpinversion,daiki2023negativepromptinversion,wallace2023edict} are proposed to invert the conditioning image into a latent embedding that achieves almost perfect reconstruction. This is then coupled with content-preserving generation techniques, usually based on the preservation of cross-attention maps with text prompts~\cite{hertz2022prompt,cao2023masactrl,parmar2023p2pzero}. While producing higher-quality visuals, these methods require additional computation and time for the optimization of the latent embeddings.

A faster alternative is to take a strictly feedforward approach~\cite{brooks2022instructpix2pix,zhang2024magicbrush,mu2024eie}, directly incorporating both conditioning image and texts in the denoising process of the diffusion model. For example, InstructPix2Pix~\cite{brooks2022instructpix2pix} generates large-scale paired data of conditioning, edited image, and editing instructions, using GPT-3~\cite{brown2020language} and Prompt-to-Prompt~\cite{hertz2022prompt}. These data are used to train a text-to-image diffusion model in a supervised way, demonstrating strong capabilities for texture transfer and object replacement. However, this type of approach is not suitable for translation tasks involving sparse conditioning signals, such as canny or edge maps. Better performance for these tasks can be achieved by approaches~\cite{zhang2023adding,li2023gligen,mou2023t2iadaptor,qin2023unicontrol,zhao2023unicontrolnet} that, like ControlNet, introduce additional trainable parameters to create a translation model. However, these parameters usually need to be trained for each type of conditioning modality, and multiple models are needed to support multiple modalities. UniControlNet~\cite{zhao2023unicontrolnet} and UniControl~\cite{qin2023unicontrol} advanced this by unifying various sparse conditions in a single model and shows promising potentials. Differently, our method jointly solve both the image editing and translation task, significantly simplifying the design and deployment of real-world applications.

\subsection{Autoregressive Models}
Autoregressive models are a class of generative models that treat text-to-image generation as a sequence-to-sequence modeling problem, similar to machine translation. Early works like VQVAE~\cite{Oord2017vqvae}, pioneered the idea of using vector quantization for this purpose, allowing  an image to be encoded as a sequence of discrete tokens, as is common for language. This enables the model to process visual data similarly to language tokens, and paved the way for a series of transformer-based autoregressive models~\cite{esser2021taming,chang2022maskgit,sun2024llamagen,chameleon,yu2022parti,ramesh2021dalle}. Recently, with the rise of large language models (LLMs)~\cite{ChatGPT,gpt3,gpt3.5,gpt4,google2023bard,devlin2019bert,igpt,anthropic2023claude,hugo2023llama,hugo2023llama2,liu2023llava,liu2023improvedllava}, LLM-style autoregressive models have started to gain popularity for text-to-image generation. Approaches like Parti~\cite{yu2022parti,chameleon,ramesh2021dalle} leverage large scale data and model sizes to successfully synthesize high quality, visually diverse images. Yet, most of these approaches remain focused on text-to-image generation, with limited efforts in adapting these models for conditional generation tasks.

Other works~\cite{chang2022maskgit,keyu2024var,zhou2024transfusion} improve autoregressive models by reconsidering the next-token prediction paradigm. They challenge the line-by-line raster-scan generation of image tokens, arguing  that images require more global contexts than text. Inspired by Masked autoencoders~\cite{he2022masked}, MaskGIT~\cite{chang2022maskgit} uses an iterative masked modeling approach. It learns to predict randomly masked tokens by attending to tokens and iteratively decodes tokens during inference. VAR~\cite{keyu2024var} changes the next-token prediction paradigm into a coarse-to-fine next-scale prediction, significantly improving the visual quality of the generated images.

The model proposed in this work builds on LlamaGen~\cite{sun2024llamagen}, which generalizes the Llama2~\cite{hugo2023llama,hugo2023llama2} architecture and is currently the state-of-the-art open-source autoregressive model for text-to-image generation. The proposed model retains its next-token prediction paradigm, leveraging LlamaGen as a strong prior for text-to-image generation, and further augments it, by incorporating both images and texts as inputs and extending its capabilities to a much wider range of conditional generation tasks.

\subsection{Combining LLM with Diffusion Models}
In image editing, there has been a recent trend to combine LLMs with diffusion models~\cite{fu2024mgie,huang2024smartedit,xia2023llmga,zhou2024transfusion}, thus addressing the limitations of CLIP-type text encoders widely used in text-to-image diffusion. These works have argued~\cite{fu2024mgie,huang2024smartedit} that replacing the CLIP~\cite{radford2021clip} text encoder with LLM modules~\cite{liu2023llava,liu2023improvedllava} enhances the reasoning capabilities of the diffusion model, enabling more nuanced interpretation of the editing instructions. However, the combination of the two models is memory, computation, and time intensive. Our aim is to design a pure autoregressive model that directly outputs the edited image as a sequence of tokens, rather than outputs in LLM text output space, thus removing the need for diffusion models. This has much lower complexity and avoids the challenges of jointly optimizing models that are trained in fundamentally different ways (autoregressive for text, denoising for images). Combined models have also only been shown successful for editing, while the proposed model also supports translation tasks involving different types of conditioning signals, e.g. the sparse signals of edge-to-image translation. This makes the proposed architecture a unified solution to a much broader range of tasks. 

\section{Method}
\label{sec:method}

\begin{figure*}
    \centering
    \includegraphics[width=0.9\linewidth]{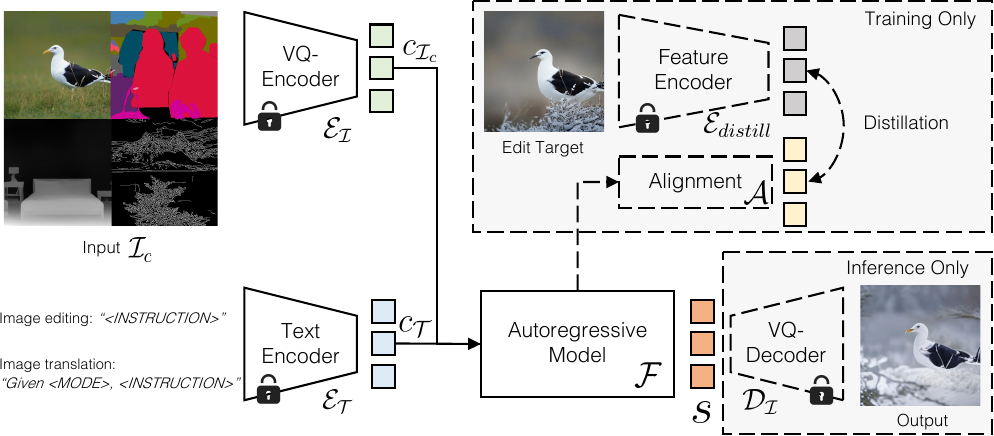}
    \vspace{-0.10in}
    \caption{Overview of EditAR, which can take various types of image conditions to perform image editing or translation. An image $\mathcal{I}_c$ is mapped through a VQ-Encoder $\mathcal{E}_{\mathcal{I}}$ to obtain corresponding token indices. Corresponding text instructions are mapped to latent embeddings $c_{\mathcal{T}}$ via a text encoder $\mathcal{E}_{\mathcal{T}}$. Both image token indices and text embeddings are input to the autoregressive transformer $\mathcal{F}$ to predict the target token indices $s$. To enhance the text-to-image alignment, a distillation loss is introduced during training to minimize the differences between the latent features of the autoregressive model, $\mathcal{F}$ and that of a feature encoder $\mathcal{E}_{distill}$. The output sequence $s$ is lastly decoded into a realistic image via a VQ-Decoder $\mathcal{D}_{\mathcal{I}}$ during inference.}
    \label{fig:overview}
    \vspace{-0.10in}
\end{figure*}

This work aims to create an autoregressive model that unifies the solution for various conditional image synthesis tasks, including different types of image editing, and image translation tasks such as edge-to-image, or segmentation-to-image. For this, we leverage existing large-scale text-to-image autoregressive models~\cite{sun2024llamagen}, which we generalize to integrate both image and text elements as conditions. Additionally, a training strategy is proposed to effectively unify various conditions, enabling high-quality image generation. 
In what follows, we first briefly review the paradigm of text-to-image autoregressive models in Section~\ref{sec:method-background}, which forms the basis of our approach. We then describe modifications, in Section~\ref{sec:method-EditAR}, to incorporate image conditioning beyond texts. Finally, we present effective learning and inference strategies for the proposed model in Section~\ref{sec:method-learning}.

\subsection{Background}
\label{sec:method-background}

Autoregressive models~\cite{esser2021taming} approach text-to-image generation as a sequence-to-sequence modeling task. A common approach includes two main components: a VQ-Autoencoder~\cite{Oord2017vqvae,razavi2019vqvae2} that converts images into discrete tokens, and an autoregressive transformer~\cite{esser2021taming} that models the categorical distribution of these tokens.

A VQ-Autoencoder uses an encoder $\mathcal{E}_\mathcal{I}$ to map an image $\mathcal{I} \in \mathcal{R}^{H \times W \times 3}$ into a latent feature map $z \in \mathcal{R}^{h \times w \times n_z}$ of feature dimensionality $n_z$. A vector quantizer is then used to map each feature vector $z_{i,j} \in \mathcal{R}^{n_z}$ into its nearest neighbor $z^q_{i,j}$ in a feature vector codebook. This allows the representation of the image as a sequence of discrete codebook indices, $s = \{s_1, s_2, \cdots, s_{h \cdot w}\}$. A decoder $\mathcal{D}_\mathcal{I}$ can finally be used to map these indices into the corresponding codebook entries, to recover the  image. 

One of the advantages of this quantization operation is that the index sequence $s$ is not fundamentally different from the sequences of one-hot codes commonly used to represent sentences in natural language. This allows a natural unified treatment of the two modalities. While large language models are autoregressive models that map a text token sequence into another text token sequence, a text-to-image generation autoregressive model outputs a sequence of visual tokens. Hence, the task of a text-to-image generation autoregressive model $\mathcal{F}$ reduces to modeling the distribution of the next index $p(s_i
|s_{<i}, c_\mathcal{T})$, where $c_\mathcal{T}$ denotes the text embeddings obtained from a text encoder $\mathcal{D}_{\mathcal{T}}$. The likelihood of the full sequence is thus defined as, 
\begin{equation}
    p(s) = \prod_{i=1}^{n}{p(s_i | s_{<i}, c_\mathcal{T})},
    \label{eq: autoregressive model}
\end{equation}
and the model parameters can be learned by maximizing the log-likelihood of the token data, $\mathbb{E}_{s \sim p(s)}[- \log p(s)]$.

\subsection{EditAR}
\label{sec:method-EditAR}

The proposed architecture for conditional image synthesis builds on LlamaGen~\cite{sun2024llamagen}, which is generalized to incorporate various types of image conditioning and a distillation loss for alignment with existing foundation models. The overall architecture is illustrated in Figure~\ref{fig:overview} and has the key design choices outlined below.

\textbf{Image Condition}.
The conditioning image $\mathcal{I}_c$ is mapped to a sequence of indices, $c_{\mathcal{I}_c} = \{c_1, c_2,  \cdots, c_{h \cdot w}\}$, via an image encoder $\mathcal{E}_{\mathcal{I}}$. The same encoder $\mathcal{E}_{\mathcal{I}}$ is also used to map the target image $\mathcal{I}_s$ into index sequence $s$. The likelihood of the output sequence $s$ becomes
\begin{equation}
    p(s) = \prod_{i=1}^{n}{p(s_i | s_{<i}, c_\mathcal{T}, c_{\mathcal{I}_c})}.
    \label{eq: image-condition autoregressive model}
\end{equation}
This is implemented with the extension of a text-to-image autoregressive model, as shown in Figure~\ref{fig:overview}. Both the additional image indices $c_{\mathcal{I}_c}$ and the text embeddings $c_\mathcal{T}$ are fed into the autoregressive model $\mathcal{F}$. During training, they are complemented by the indices $s$ that appear at both input $s_{<i}$ and output $s_i$ of the model to implement the autoregressive operation. Note that different sets of positional embeddings are applied to the embeddings of $c_{\mathcal{I}_c}$ and $s$ to differentiate the control image sequence from the output sequence. In addition, an unconditional image embedding is introduced and used during training to preserve the model's text-to-image and unconditional generation ability.


\textbf{Incorporating Image Modalities}. 
Our method supports various imaging modalities, including canny edges, depth maps, segmentation masks, and natural images. The generation process is adjusted to these different modalities by modifying the phrasing of text inputs. For instance, to generate an image from a depth map, we use the ``\textit{Given the depth, generate the image following the instruction:} {\tt<INSTRUCTION>}'', where the instruction is usually the description of the content of the generated image. This approach is similarly applied for canny edge maps and segmentation masks.  When the input is a real image, we use only ``{\tt <INSTRUCTION>}'' to specify how to modify the input image.

\textbf{Distillation}. 
Distillation from vision foundation models has been demonstrated to be effective across a variety of computer vision tasks. An autoregressive model gradually learns to synthesize tokens of high likelihood given the conditioning token stream. However, the autoregressive model, trained solely to predict token indices, are not guaranteed to learn general semantic features. To inject general visual knowledge in the feature space, we introduce a distillation loss from the vision foundation model feature encoder $\mathcal{E}_{distill}$, which in our implementation is DINOv2~\cite{caron2021dino,oquab2024dinov2}. An alignment network $\mathcal{A}$, composed of a single convolutional layer, is used to match the dimensionality of the embedding space of the autoregressive model $\mathcal{F}(\cdot)$ with that of the foundation model. During training, the parameters of this network are learned to minimize the distillation loss
\begin{equation}
    \mathcal{L}_{distill}=MSE \Big( \mathcal{A}(\mathcal{F}(\cdot)), \mathcal{E}_{distill}(\cdot) \Big).
\end{equation}
For both $\mathcal{F}$ and $\mathcal{E}_{distill}$, the features extracted from the last hidden layer are used to compute this loss.  
Empirically, we find this design to improve the text-to-image alignment.

\subsection{Training and Inference}
\label{sec:method-learning}

During training, both the text-encoder $\mathcal{E}_{\mathcal{T}}$ and foundation model $\mathcal{E}_{distill}$ are frozen. The parameters of the autoregressive transformer $\mathcal{F}$ are initialized from pre-trained text-to-image models. All parameters of $\mathcal{F}$ and $\mathcal{A}$ are optimized for the adaptation, using
\begin{equation}
    \mathcal{L} = \mathcal{L}_{CE} + \lambda_{distill} \cdot \mathcal{L}_{distill},
    \label{eq:loss}
\end{equation}
where $\mathcal{L}_{CE}=\mathbb{E}_{x \sim p(x)}[- \log p(s)]$ is the cross-entropy loss commonly used to train next-token prediction models. 

Motivated by prior works~\cite{brooks2022instructpix2pix,sun2024llamagen}, we apply dropout during training to preserve the model's unconditional generation ability, and classifier-free guidance during inference. For this, we set $c_{\mathcal{T}} = \emptyset$ in $5\%$ of the training examples, $c_{\mathcal{I}_c} = \emptyset$ in another $5\%$, and both $c_{\mathcal{T}} = \emptyset$ and $c_{\mathcal{I}_c} = \emptyset$ in a final $5\%$. At inference, only $c_{\mathcal{I}_c}$ and  $c_{\mathcal{T}}$ are given as inputs, and the set $s$ is predicted sequentially. Classifier-free guidance consist of using 
\begin{equation}
    \begin{aligned}
        \log &p(s_i | s_{<i}, c_\mathcal{T}, c_{\mathcal{I}}) = \log p(s_i| s_{<i}, c_{\mathcal{I}}) \\ &+ \eta \cdot \Big( \log p(s_i| s_{<i}, c_\mathcal{T}, c_{\mathcal{I}}) - \log p(s_i| s_{<i}, c_{\mathcal{I}}) \Big),
        \label{eq: cfg}
    \end{aligned}
\end{equation}
where $\eta$ is a guidance strength hyperparameter, to compute index log-probabilities at inference.

\section{Experiments}
\label{sec:experiments}
In this section, we discuss an extensive experimental evaluation of EditAR on various conditional image synthesis tasks. Note that EditAR is a single model that solves both image editing, as discussed in Section ~\ref{sec:experiments-image editing}, and image translation, as discussed in Section~\ref{sec:experiments-image translation}, while all baseline models we compare to are specialized models for a subset of tasks. Surprisingly, despite this disadvantage, EditAR still shows strong performance.

\begin{figure*}
    \centering
    \includegraphics[width=1.0\linewidth]{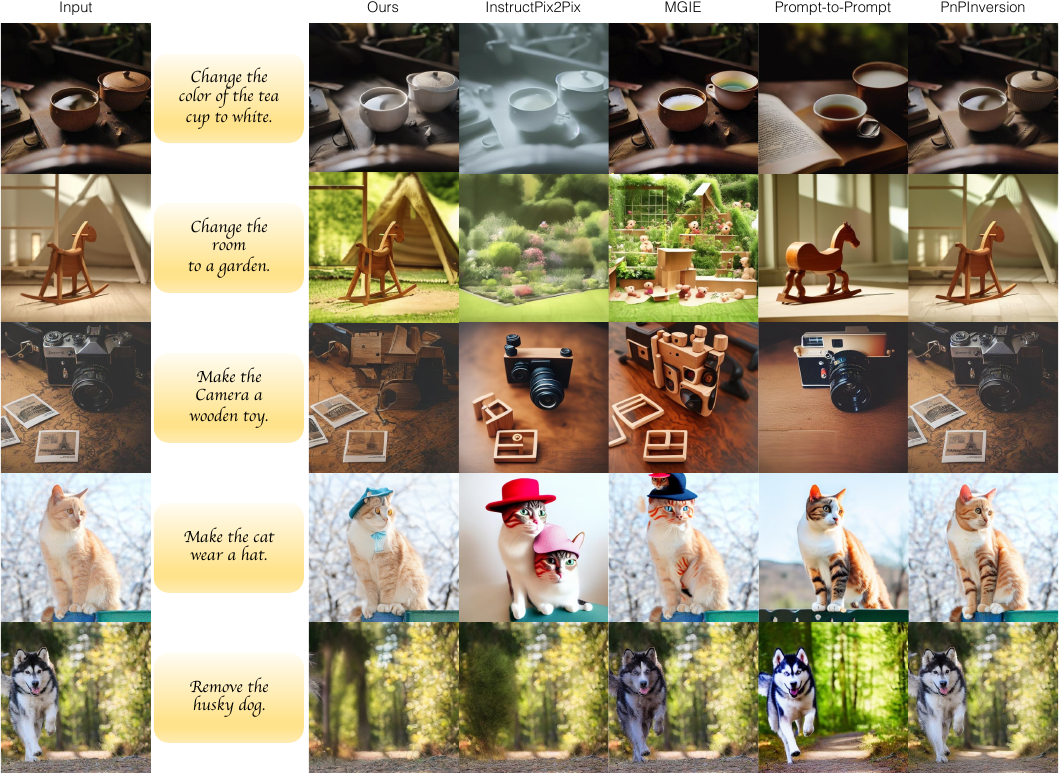}
    \vspace{-0.10in}
    \caption{Comparison of EditAR (Ours) to feed-forward methods (InstructPix2Pix~\cite{brooks2022instructpix2pix}, MGIE~\cite{fu2024mgie}) and inversion-based approaches (Prompt-to-Prompt~\cite{hertz2022prompt}, PnPInversion~\cite{ju2023pnpinversion}) on various edits. Our method preserves input details well and has strong text-to-image alignment. In contrast, baseline results exhibit unrealistic visuals, including exaggerated edits or not following instructions, unrealistic modifications, or are unable to localize objects accurately.}
    \label{fig:image-editing}
    \vspace{-0.10in}
\end{figure*}

\subsection{Dataset and Training Details}
\label{sec:experiments-dataset}

\textbf{Dataset}. EditAR is trained in a fully supervised manner using paired data. Recent works~\cite{brooks2022instructpix2pix,hui2024hqedit,ge2024seedxedit} have shown that it is possible to create large-scale image editing data with automated pipelines. We use 1.5M examples from SEED-Data-Edit-Unsplash~\cite{ge2024seedxedit}, created with ChatGPT~\cite{ChatGPT} and Plug-and-play~\cite{tumanyan2023plugandplay}, for a range of image editing tasks, including modifying styles, objects, colors, and materials. To further support editing operations like object addition and removal, we add the PIPE dataset~\cite{wasserman2024paint} with 1.8M examples, where image pairs for the task of adding objects are created by inpainting. During training, we randomly flip each pair with a $50\%$ probability and adjust the editing instruction from ``Add'' to ``Remove'' correspondingly. For image translation tasks, we follow ControlNet++~\cite{li2024controlnetplusplus} and use COCOStuff~\cite{caesar2018cocostuff} for segmentation mask-to-image translation and MultiGen-20M for canny edge and depth-to-image tasks. 

\textbf{Evaluation and Metrics}.
To systematically evaluate EditAR, we use multiple benchmark datasets. For image editing, we use the PIE-Bench dataset~\cite{ju2023pnpinversion} with $700$ examples, covering 10 editing types. Our method uses the source image and editing instructions to predict the target edit. Both reconstruction and text-to-image alignment are evaluated as in~\cite{ju2023pnpinversion} with the annotated foreground masks. For the evaluation of image translation, we follow ControlNet++~\cite{li2024controlnetplusplus} and use the corresponding validation splits for COCOStuff~\cite{caesar2018cocostuff} and MultiGen-20M~\cite{qin2023unicontrol}, which contain $5,000$ examples per task. Regarding metrics, we follow the common practice in the field: mIOU is used for semantic segmentation conditions, RMSE for depth map conditions, and SSIM for canny edge conditions. For canny edge condition, (100, 200) are used as thresholds.

\textbf{Training and Inference}.
The training hyperparameters follow the setting of~\cite{sun2024llamagen}. All images are resized to a resolution of $512 \times 512$ for both training and inference. The VQ-Autoencoder has a downsampling ratio of 16, so that each image is represented by 1024 tokens. The model is optimized using AdamW with a constant learning rate of $10^{-4}$, $\beta_1 = 0.9$, $\beta_2 = 0.95$, and weight decay of $0.05$. The model is trained with a batch size of 64 for $40,000$ iterations. In all experiments, we use $\lambda_{distill}=0.5$ in (\ref{eq:loss}). For the guidance strength of~(\ref{eq: cfg}), we empirically find that $\eta = 3.0$ yields a good balance between reconstruction quality and text-to-image alignment.  More details and results are presented in the supplementary materials.

\begin{table*}[t!]
\centering 
\begin{tabular}{c|c|c|c|c|c|c|c|c}
\toprule[2pt]
\multicolumn{1}{c|}{} & \multicolumn{1}{c|}{} & \multicolumn{1}{c|}{\textbf{Structure}} & \multicolumn{4}{c|}{\textbf{Background Preservation}} & \multicolumn{2}{c}{\textbf{CLIP Similarity}} \\ \cline{3-9} 
\multicolumn{1}{c|}{\multirow{-2}{*}{\textbf{Method}}} & {\multirow{-2}{*}{\textbf{\begin{tabular}[c]{@{}c@{}}T2I\\Model\end{tabular}}}} & \multicolumn{1}{c|}{\textbf{Distance} $\downarrow$} & \multicolumn{1}{c|}{\textbf{PSNR} $\uparrow$} & \multicolumn{1}{c|}{\textbf{LPIPS} $\downarrow$} & \multicolumn{1}{c|}{\textbf{MSE} $\downarrow$} & \multicolumn{1}{c|}{\textbf{SSIM} $\uparrow$} & \multicolumn{1}{c}{\textbf{Whole} $\uparrow$} & \multicolumn{1}{c}{\textbf{Edited} $\uparrow$}\\ \cline {1-9}
Prompt-to-Prompt & SD1.4 & 69.43 & 17.87 & 208.80 & 219.88 & 71.14 & 25.01 & 22.44 \\
Null-text Inversion& SD1.4 & 13.44 & 27.03 & 60.67 & 35.86 & 84.11 & 24.75 & 21.86 \\
PnPInversion & SD1.4 & 11.65 & 27.22 & 54.55 & 32.86 & 84.76 & 25.02 & 22.10 \\
Pix2pix-zero & SD1.4 & 61.68 & 20.44 & 172.22 & 144.12 & 74.67 & 22.80 & 20.54 \\
MasaCtrl & SD1.4 & 28.38 & 22.17 & 106.62 & 86.97 & 79.67 & 23.96 & 21.16 \\
\bottomrule
InstructPix2Pix & SD1.5 & 107.43 & 16.69 & 271.33 & 392.22 & 68.39 & 23.49 & \textbf{22.20} \\
MGIE & SD1.5 & 67.41 & 21.20 & 142.25 & 295.11 & \textbf{77.52} & 24.28 & 21.79 \\
EditAR (Ours) & LlamaGen &  \textbf{39.43} & \textbf{21.32} & \textbf{117.15} & \textbf{130.27} & 75.13 & \textbf{24.87} & 21.87 
\\ \bottomrule[2pt]
\end{tabular}%
\caption{Comparison of EditAR to various feed-forward methods (bottom) and inversion-based approaches (top) on the PIE-Bench dataset~\cite{ju2023pnpinversion}. Our method achieves the highest overall performance among all feed-forward approaches and narrows the gap with advanced inversion-based methods. While InstructPix2Pix~\cite{brooks2022instructpix2pix} achieves a high edited CLIP score, it struggles to reconstruct unedited regions accurately, as indicated by its lower whole CLIP score and background scores. MGIE~\cite{fu2024mgie} shows improved background preservation and editing quality, yet our method demonstrates stronger overall performance. }
\label{tab:image-editing}
\end{table*}

\begin{table*}[t!]
\centering 
\begin{tabular}{c|c|c|c|c|c|c|c}
\toprule[2pt]
\multicolumn{1}{c|}{} & \multicolumn{1}{c|}{} &  \multicolumn{2}{c|}{\textbf{Depth Map (MultiGen)}} & \multicolumn{2}{c|}{\textbf{Canny Edge (MultiGen)}} & \multicolumn{2}{c}{\textbf{Seg. Mask (COCOStuff)}}  \\ \cline{3-8} 
\multicolumn{1}{c|}{\multirow{-2}{*}{\textbf{Method}}} & {\multirow{-2}{*}{\textbf{\begin{tabular}[c]{@{}c@{}}T2I\\Model\end{tabular}}}} & \multicolumn{1}{c|}{\textbf{RMSE} $\downarrow$} & \multicolumn{1}{c|}{\textbf{FID} $\downarrow$} & \multicolumn{1}{c|}{\textbf{SSIM} $\uparrow$} & \multicolumn{1}{c|}{\textbf{FID} $\downarrow$} & \multicolumn{1}{c|}{\textbf{mIOU} $\uparrow$} & \multicolumn{1}{c}{\textbf{FID} $\downarrow$} \\ \cline {1-8}
T2I-Adapter & SD1.5 & 48.40 & 22.52 & 38.93 & 15.96 & - & -  \\
UniControlNet & SD1.5 & 40.65 & 22.27 & 41.55 & 17.14 & - & -  \\
UniControl & SD1.5 & 39.18 & 18.66 & 51.71 & 19.94 & - & -   \\
ControlNet & SD1.5 & 35.90 & 17.76 & 54.87 & 14.73 & 27.46 & 21.33   \\
ControlNet++ & SD1.5 & \textbf{28.32} & 16.66 & 57.06 & 18.23 & \textbf{34.56} & 19.29   \\
EditAR (Ours) & LlamaGen & 34.93 & \textbf{15.97} & 48.11 & \textbf{13.91} & 22.62 & \textbf{16.13} 
\\ \bottomrule[2pt]
\end{tabular}%
\caption{Comparison of EditAR to various conditional image synthesis baselines. Our method yields the best FID scores, underscoring its strength in both sample quality and diversity. Note that T2I-Adaptor~\cite{mou2023t2iadaptor}, ControlNet~\cite{zhang2023adding}, and ControlNet++~\cite{li2024controlnetplusplus} train separate models for different conditions. While UniControlNet~\cite{zhao2023unicontrolnet} and UniControl~\cite{qin2023unicontrol} show promise as unified models capable of handling multiple tasks, they underperform EditAR. All the results are compared on $512 \times 512$ image resolution with
Clean-FID implementation~\cite{parmar2022cleanfid}.
}
\label{tab:image-translation}
\end{table*}

\begin{figure*}
    \centering
    \includegraphics[width=1.0\linewidth]{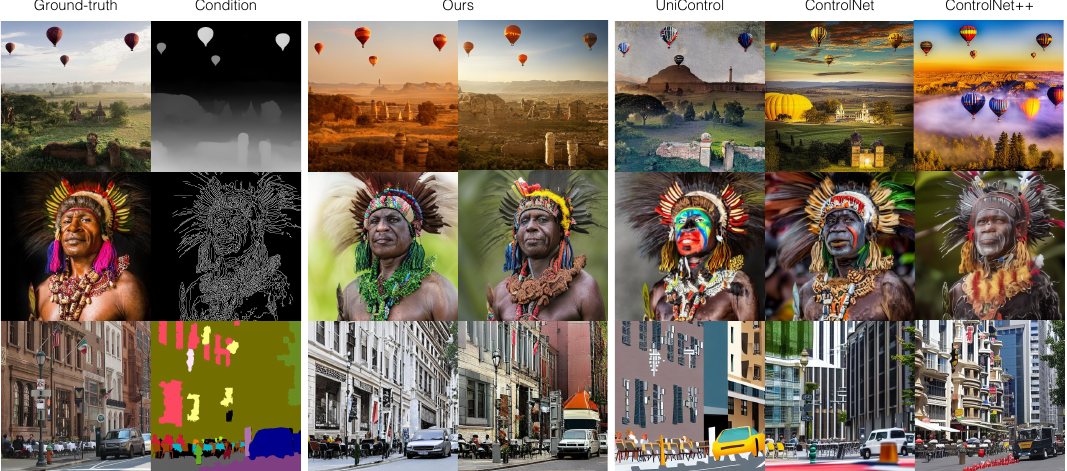}
    \vspace{-0.20in}
    \caption{Visual comparisons to baseline methods on various image translation tasks. Our method, EditAR, produces photo-realistic results, preserves input details, and offers substantial sample diversity.}
    \label{fig:image-translation}
    \vspace{-0.20in}
\end{figure*}

\subsection{Image Editing}
\label{sec:experiments-image editing}

For image editing, we compare EditAR to various baselines, as shown in Table~\ref{tab:image-editing} and Figure~\ref{fig:image-editing}. Specifically, we consider two types of diffusion baselines: feed-forward instruction-based methods and inversion-based content-preserving editing methods.

EditAR achieves the best overall performance among feed-forward instruction-based baselines, e.g., InstructPix2Pix~\cite{brooks2022instructpix2pix} and MGIE~\cite{fu2024mgie}. All methods share the same inputs: an input image and an instruction prompt. InstructPix2Pix excels particularly in texture transfer and object replacement tasks. MGIE leverages both multimodal large language models (MLLMs) to generate expressive instructions, which provides explicit guidance to enhance the editing process. However, as shown in Table~\ref{tab:image-editing} and Figure~\ref{fig:image-editing}, while these methods achieve great text-to-image alignment, they often struggle to preserve the background or produce exaggerated image edits. In comparison, EditAR maintains a good balance between reconstruction and editing quality, resulting in highly photorealistic images.

Unlike feed-forward methods, inversion-based approaches first invert an image into a latent space, usually latent noise or embeddings, before performing content-preserving sampling to generate the edited target. Instead of directly taking an editing instruction, these methods additionally require a source prompt, which is crucial for the inversion process. As shown in Table~\ref{tab:image-editing}, though Prompt-to-Prompt yields the strong editing clip scores, they fail to preserve the background (see also Figure~\ref{fig:image-editing}). 
With further optimization, approaches like Null-text Inversion~\cite{mokady2023null} and PnpInversion~\cite{ju2023pnpinversion} show improved reconstruction fidelity. However, they remain limited by the absence of a unified, content-preserving model that performs consistently well across diverse tasks.

\begin{figure}
    \centering
    \includegraphics[width=1.0\linewidth]{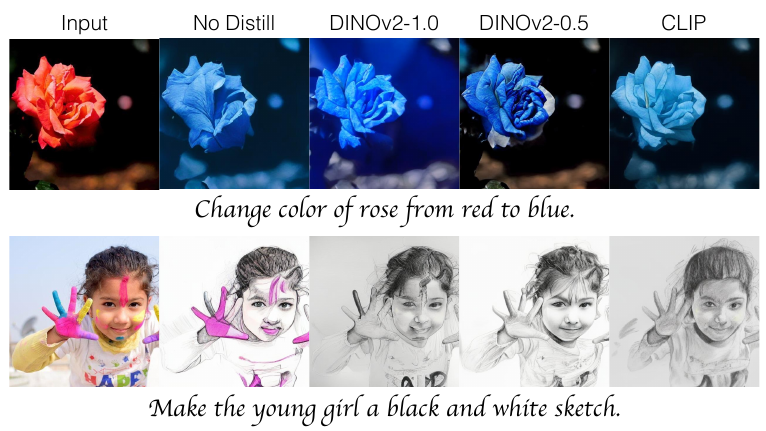}
    \vspace{-0.20in}
    \caption{Studies on distillation loss. From left to right, we show the input image, results w/out distillation, and distillation results with DINOv2 and CLIP. The top example shows improved object localization. The bottom shows better text-to-image alignment. }
    \label{fig:ablation-image-editing}
    \vspace{-0.10in}
\end{figure}

\begin{figure}
    \centering
    \includegraphics[width=1.0\linewidth]{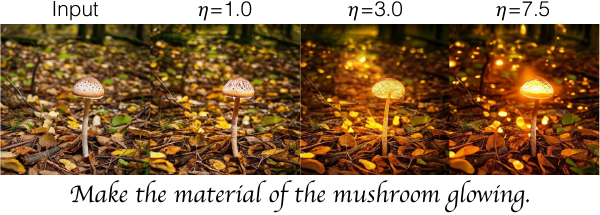}
    \vspace{-0.20in}
    \caption{Ablation of classifier-free guidance values. }
    \label{fig:ablation-cfg}
    \vspace{-0.10in}
\end{figure}

\subsection{Image Translation}
\label{sec:experiments-image translation}

We validate the effectiveness of the proposed method on three image translation tasks: segmentation-to-image, edge-to-image, and depth-to-image. As shown in Table~\ref{tab:image-translation} and Figure~\ref{fig:image-translation}, EditAR outperforms other methods in FID across all tasks, demonstrating its ability to synthesize images that are both high-quality and diverse.

Compared with unified-model approaches (UniControl~\cite{qin2023unicontrol} and UniControlNet~\cite{zhao2023unicontrolnet}), our method demonstrates superior scores. As evidenced by both approaches, unifying various conditional synthesis tasks alone is challenging. Specifically, UniControl employs a task-aware hyper-network to modulate the zero-convolution modules of ControlNet. UniControlNet fine-tunes two additional adapters on frozen pre-trained text-to-image diffusion models, enabling various control inputs. Despite these advances, EditAR still delivers better overall performance.

Single-modality approaches take advantage of task-specific fine-tuning to achieve optimal performance on each domain. For example, in the segmentation-to-image task, single-model approaches often generate pre-defined classes, as they are trained solely to generate these specific categories. While EditAR consistently produces visually appealing results, the semantics may not always align perfectly with the ground-truth. Though learning a more challenging task, our model still synthesizes diverse images with good visual quality.

\subsection{Ablation}
\label{sec:experiments-ablation}
In this section, we study the effectiveness of the distillation loss, and show results distilled from different foundation models. Furthermore, we also explore the effects of different classifier-free-guidance, and show that it is crucial for high synthesis quality.

\textbf{Distillation.}
We hypothesize that an autoregressive model trained solely to predict token indices is not guaranteed to align with semantic features. Therefore, we propose a distillation loss to encourage stronger feature space similarity between the autoregressive model and foundation models. As shown in Figure~\ref{fig:ablation-image-editing}, experimental results show that adding the distillation loss improves the overall text-to-image alignment, e.g., better localizing the target editing object. We try different foundation models and empirically find that DINOv2-0.5, with distillation coefficient $\lambda_{distill}=0.5$, achieves the best overall performance.

\textbf{CFG Guidance.}
We experiment with various guidance coefficients, as shown in Figure~\ref{fig:ablation-cfg}. Our results indicate that an optimal classifier-guidance strength is crucial for achieving high visual quality. With a lower classifier-free guidance value, the model shows weak responses to the text input. While a higher value improves text-to-image alignment, the reconstruction quality is reduced. In practice, we find that a value of $\eta = 3$ yields the best trade-off between text-to-image alignment and reconstruction quality.

\section{Conclusion}

We introduce EditAR, a unified autoregressive framework designed for a wide range of conditional image generation tasks. EditAR adapts to various image inputs, switching modes solely through text prompts. We assess its effectiveness across diverse image editing and translation tasks, demonstrating its strong performance of reconstruction quality and edited visual quality. As text-to-image autoregressive models continue to advance, we hope this work paves the way for new possibilities in unified conditional generative modeling.

\vspace{1em}
{\footnotesize \textbf{Acknowledgements.}~ This project was supported, in part, by NSF CAREER Award IIS-2240014, NSF award IIS-2303153, and gifts from Qualcomm and Meta. We also acknowledge and thank the use of the Nautilus platform for the experiments discussed above.}

{
    \small
    \bibliographystyle{ieeenat_fullname}
    \bibliography{main}
}
\newpage
\appendix

In this supplementary materials, we provide more details of the submission. We show additional editing results and more baselines in Section~\ref{supp section: additional image editing comparison} complementing Section 4.2 in the paper; Furthermore, more image translation and comparisons are presented in Section~\ref{supp section: additional image translation comparison}; More results of image translation with distillation are discussed in Section~\ref{supp section: additional distillation comparison}. More implementation details and training recipes (paper Section 4.1) are discussed in Section~\ref{supp section: implementation details}. \textit{The models and codes will be released.}

\begin{table*}[t!]
\centering 
\begin{tabular}{c|c|c|c|c|c|c|c|c}
\toprule[2pt]
\multicolumn{1}{c|}{} & \multicolumn{1}{c|}{} & \multicolumn{1}{c|}{\textbf{Structure}} & \multicolumn{4}{c|}{\textbf{Background Preservation}} & \multicolumn{2}{c}{\textbf{CLIP Similarity}} \\ \cline{3-9} 
\multicolumn{1}{c|}{\multirow{-2}{*}{\textbf{Method}}} & {\multirow{-2}{*}{\textbf{\begin{tabular}[c]{@{}c@{}}T2I\\Model\end{tabular}}}} & \multicolumn{1}{c|}{\textbf{Distance} $\downarrow$} & \multicolumn{1}{c|}{\textbf{PSNR} $\uparrow$} & \multicolumn{1}{c|}{\textbf{LPIPS} $\downarrow$} & \multicolumn{1}{c|}{\textbf{MSE} $\downarrow$} & \multicolumn{1}{c|}{\textbf{SSIM} $\uparrow$} & \multicolumn{1}{c}{\textbf{Whole} $\uparrow$} & \multicolumn{1}{c}{\textbf{Edited} $\uparrow$}\\ \cline {1-9}
Prompt-to-Prompt & SD1.4 & 69.43 & 17.87 & 208.80 & 219.88 & 71.14 & 25.01 & 22.44 \\
Null-text Inversion& SD1.4 & 13.44 & 27.03 & 60.67 & 35.86 & 84.11 & 24.75 & 21.86 \\
PnP Inversion & SD1.4 & 11.65 & 27.22 & 54.55 & 32.86 & 84.76 & 25.02 & 22.10 \\
Pix2Pix-Zero & SD1.4 & 61.68 & 20.44 & 172.22 & 144.12 & 74.67 & 22.80 & 20.54 \\
MasaCtrl & SD1.4 & 28.38 & 22.17 & 106.62 & 86.97 & 79.67 & 23.96 & 21.16 \\
\bottomrule
InstructPix2Pix & SD1.5 & 107.43 & 16.69 & 271.33 & 392.22 & 68.39 & 23.49 & 22.20 \\
MagicBrush & SD1.5 & 26.81 & 26.85 & 66.67 & 171.11 & 83.37 & 23.89 & 20.84 \\
InstructDiffusion & SD1.5 & 74.21 & 20.88 & 142.35 & 353.45 & 76.70 & 24.06 & 21.57 \\
MGIE & SD1.5 & 67.41 & 21.20 & 142.25 & 295.11 & 77.52 & 24.28 & 21.79 \\
SEED-X-Edit & SD-XL & 61.69 & 18.80 & 173.63 & 209.05 & 74.93 & 25.51 & 22.20 \\
EditAR (Ours) & LlamaGen & 39.43 & 21.32 & 117.15 & 130.27 & 75.13 & 24.87 & 21.87 
\\ \bottomrule[2pt]
\end{tabular}%
\caption{Comparisons complementing Table 1. Comparison of EditAR to various feed-forward methods (bottom) and inversion-based approaches (top) on the PIE-Bench dataset. Our results attain superior results in preserving the details of the input as well as following the new edits, narrowing the gap with advanced inversion-based methods. The feed-forward baseline results show various types of failures, such as decline in image quality, unfaithful background preservation, and not following the editing instructions. While InstructPix2Pix achieves a high edited CLIP score, it struggles to reconstruct unedited regions accurately, as indicated by its lower whole CLIP score and background scores. MagicBrush shows improved background preservation but at the expense of editing quality. InstructDiffusion and MGIE shows improved reconstruction and editing quality, yet our method demonstrates stronger overall performance. Seed-X-Edit struggles with reconstructing unedited regions and produces images with unrealistic contrast.
 }
\label{supp tab: image editing}
\end{table*}

\section{Additional Image Editing Comparison} \label{supp section: additional image editing comparison}

\begin{figure*}[t]
    \centering
    \includegraphics[width=0.9\linewidth]{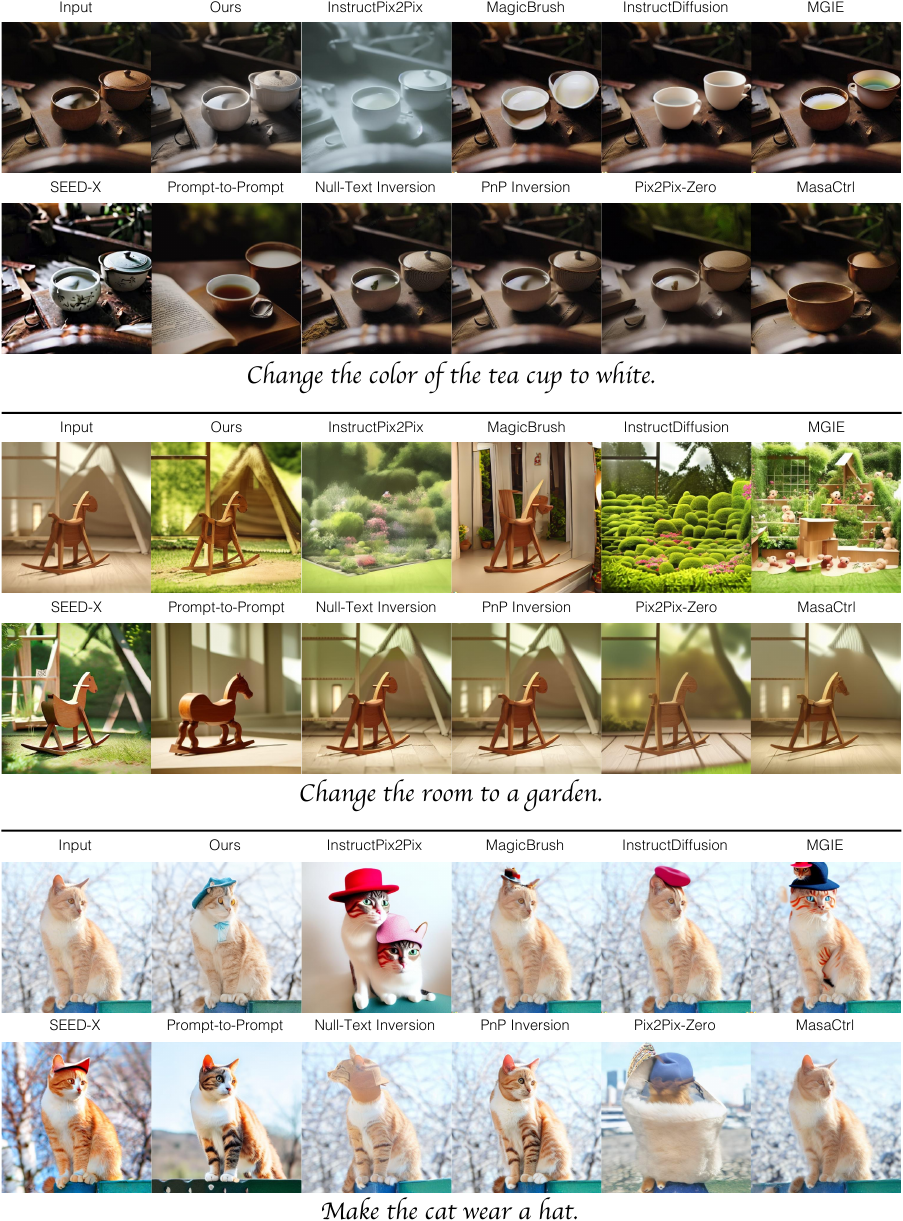}
    \caption{Comparisons complementing Figure 3. Comparison of EditAR to various feed-forward methods  and inversion-based approaches on the PIE-Bench dataset. Our results attain superior results in preserving the details of the input as well as following the given edits.}
    \label{supp fig: image editing 1}
\end{figure*}

\begin{figure*}[t]
    \centering
    \includegraphics[width=0.9\linewidth]{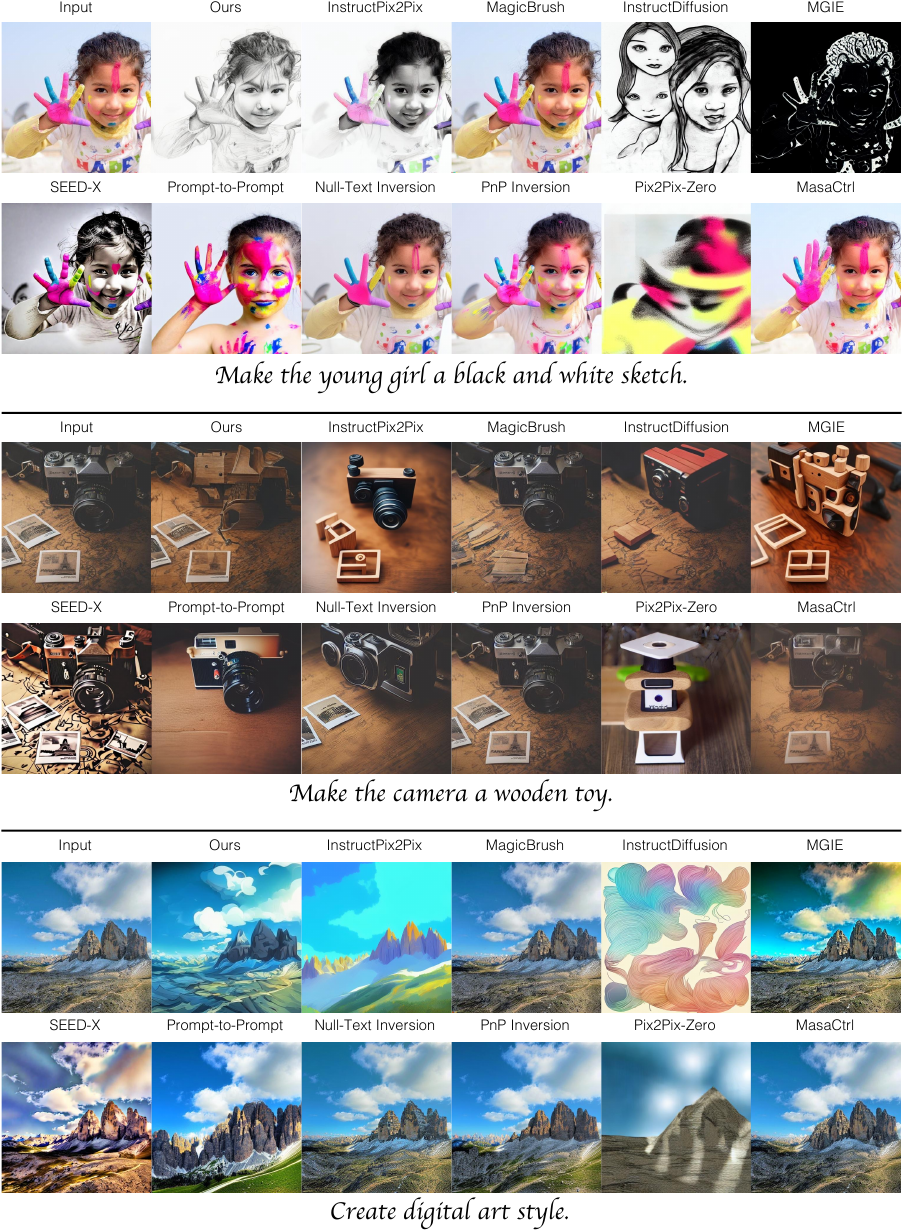}
    \caption{Comparisons complementing Figure 3. Comparison of EditAR to various feed-forward methods  and inversion-based approaches on the PIE-Bench dataset. Our results attain superior results in preserving the details of the input as well as following the given edits.}
    \label{supp fig: image editing 2}
\end{figure*}

\begin{figure*}[t]
    \centering
    \includegraphics[width=0.9\linewidth]{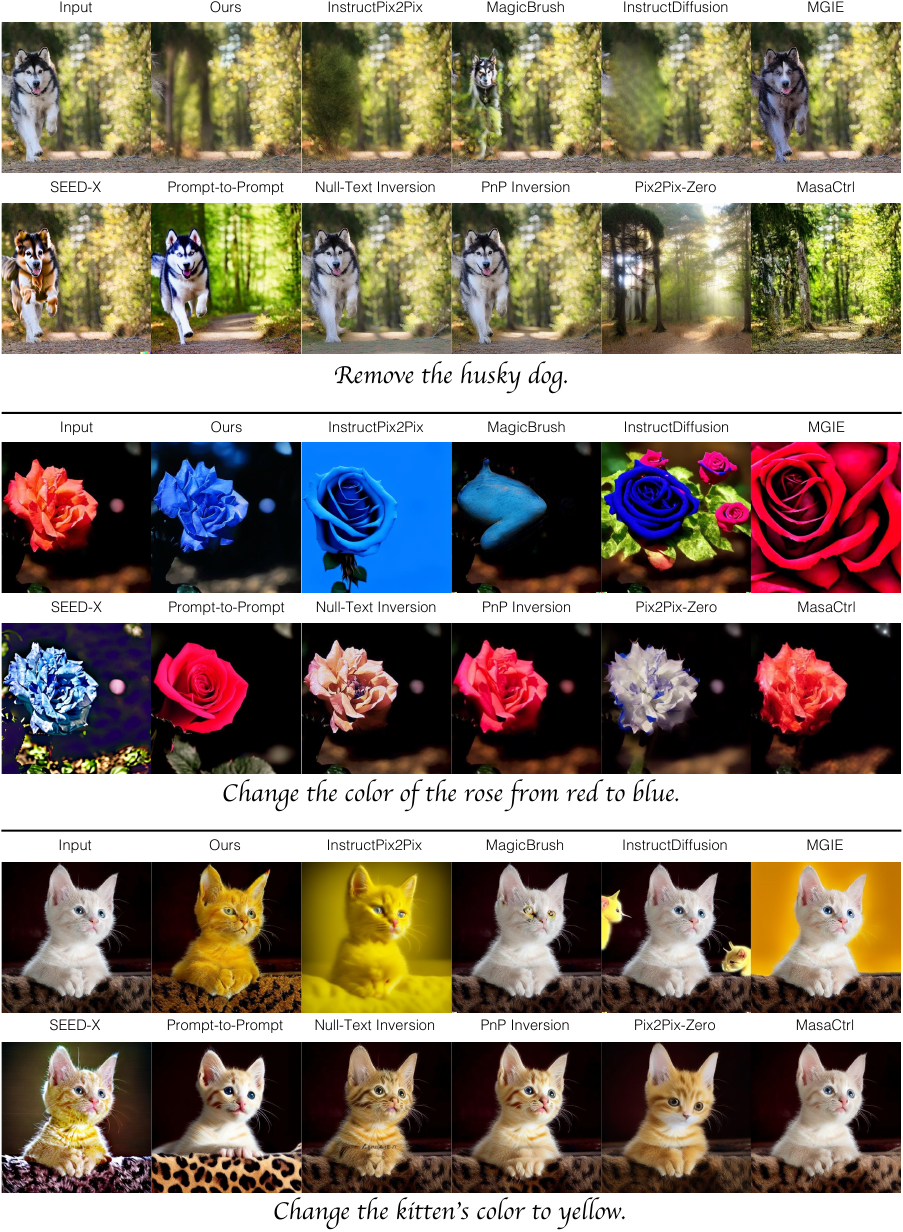}
    \caption{Comparisons complementing Figure 3. Comparison of EditAR to various feed-forward methods  and inversion-based approaches on the PIE-Bench dataset. Our results attain superior results in preserving the details of the input as well as following the given edits.}
    \label{supp fig: image editing 3}
\end{figure*}

\begin{figure*}[t]
    \centering
    \includegraphics[width=0.9\linewidth]{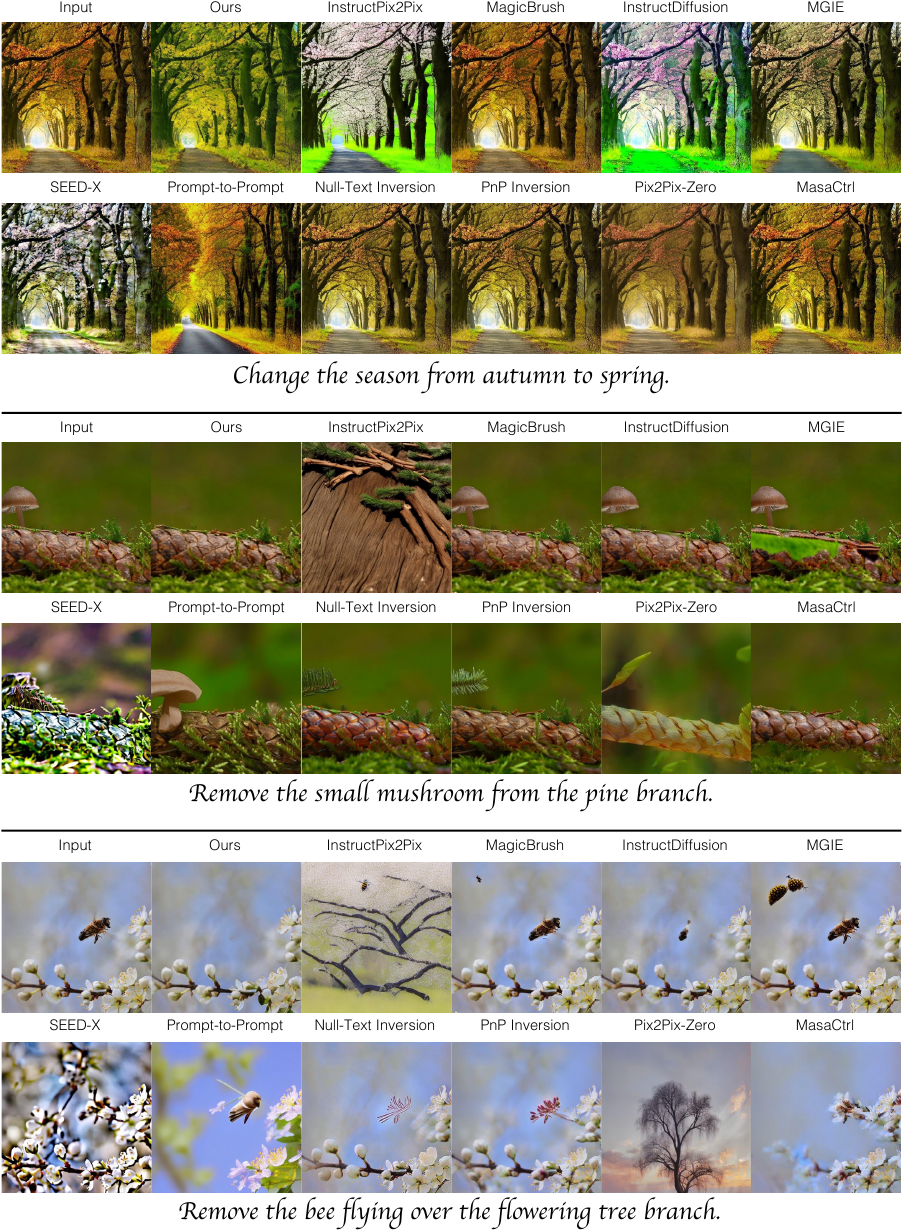}
    \caption{Comparisons complementing Figure 3. Comparison of EditAR to various feed-forward methods  and inversion-based approaches on the PIE-Bench dataset. Our results attain superior results in preserving the details of the input as well as following the given edits.}
    \label{supp fig: image editing 4}
\end{figure*}

\begin{figure*}[t]
    \centering
    \includegraphics[width=0.9\linewidth]{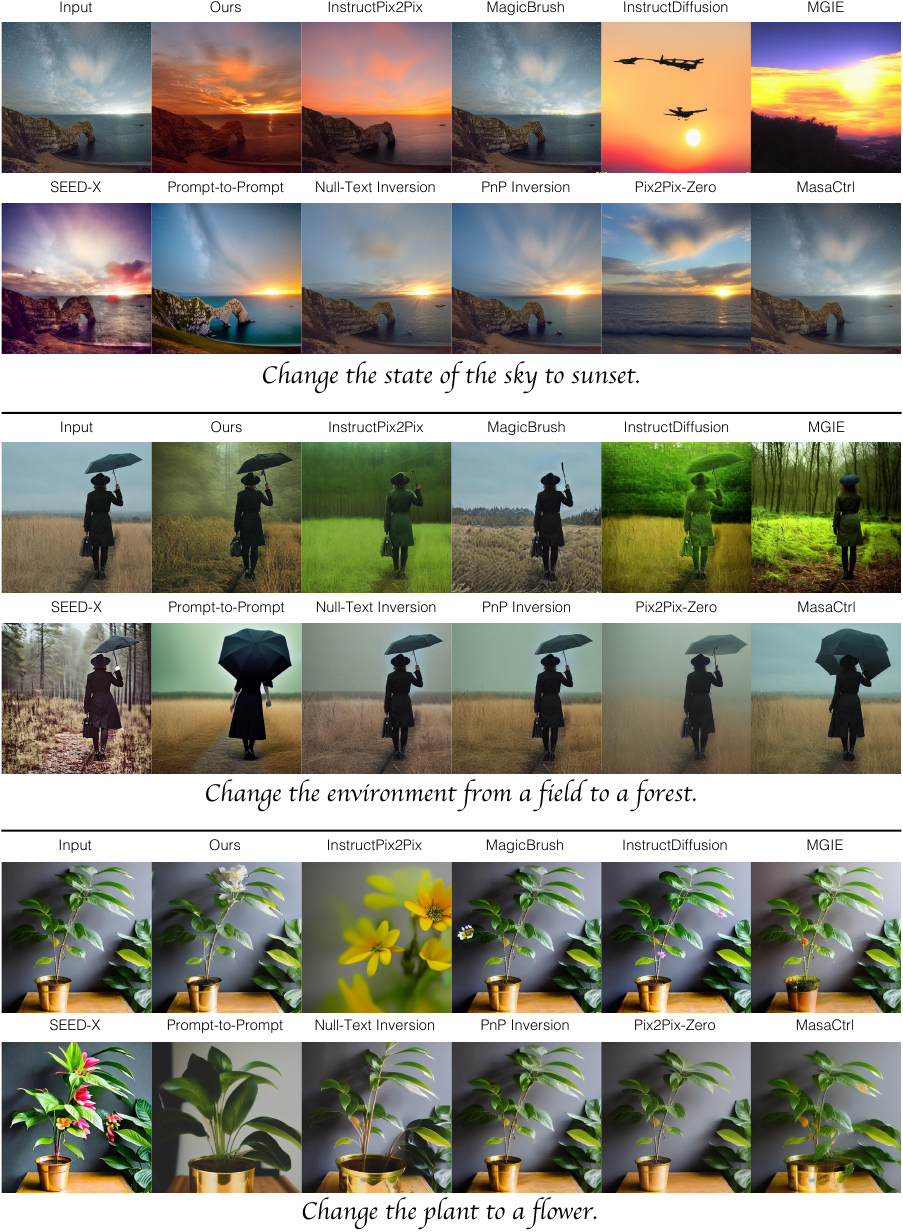}
    \caption{Comparisons complementing Figure 3. Comparison of EditAR to various feed-forward methods  and inversion-based approaches on the PIE-Bench dataset. Our results attain superior results in preserving the details of the input as well as following the given edits.}
    \label{supp fig: image editing 5}
\end{figure*}

In Table 1, Figure 3 and Section 4.2 in the main text, we have shown our editing results as well as comparisons to various baselines. Here we provide more details and show additional comparisons to more baselines, as shown in Table~\ref{supp tab: image editing}. More visual comparisons are presented in Figure~\ref{supp fig: image editing 1}, ~\ref{supp fig: image editing 2}, ~\ref{supp fig: image editing 3}, ~\ref{supp fig: image editing 4}, ~\ref{supp fig: image editing 5}. For all methods, we use their officially released model checkpoints to ensure quality. Note that in the paper where we referenced PnP Inversion, the method is also referred to as Direct Inversion. For consistency, we use PnP Inversion in the paper. We elaborate on the detailed implementations of each baseline below.

\textbf{InstructPix2Pix}. Instructional image editing poses greater challenges compared to text-to-image generation, as it usually requires the ability to process images following instructions while maintaining visual realism and reconstruction quality. InstructPix2Pix pioneered instruction-based image editing by creating a large-scale dataset comprising conditioning images, edited outputs, and corresponding editing instructions, then training a text-to-image diffusion model in a fully supervised manner. This method excels in tasks such as global texture transfer and object replacement. However, while InstructPix2Pix achieves a high edited CLIP score, it struggles to accurately reconstruct unedited regions, as observed by its lower whole CLIP score and background preservation scores. As visualized in Figure~\ref{supp fig: image editing 1}, ~\ref{supp fig: image editing 2}, ~\ref{supp fig: image editing 3}, ~\ref{supp fig: image editing 4}, ~\ref{supp fig: image editing 5}, results often reveal exaggerated edits and unrealistic modifications. To further evaluate if the differences are introduced by data alone, we fine-tuned InstructPix2Pix on the same datasets (SEED-Data-Edit-Unsplash and PIPE Dataset) as our proposed method. Results show that the model only achieves an editing CLIP score of 19.97, with most examples failing to follow the given instructions. This shows the method is highly sensitive to input data, making it challenging to balance multiple datasets effectively.

\textbf{MagicBrush}. To enhance the editing quality of InstructPix2Pix, MagicBrush introduces a manually annotated dataset of 10K real image pairs (source image, instruction, target image) across diverse editing scenarios. As shown in Table~\ref{supp tab: image editing}, fine-tuning InstructPix2Pix on the MagicBrush dataset enhances reconstruction performance but leads to a significant decline in edited CLIP scores, highlighting the sensitivity of balancing data. Moreover, manual dataset annotation is time-consuming and challenging to scale efficiently.

\textbf{InstructDiffusion}. InstructDiffusion aims to develop a unified model capable of addressing a wide range of vision tasks without requiring task-specific modifications. To enable this, it extends InstructPix2Pix by training on diverse tasks, including understanding tasks (e.g., segmentation and keypoint detection) and generative tasks (e.g., editing and enhancement). The approach consists of two stages: training a unified model across various tasks, followed by fine-tuning for specific tasks, similar to InstructPix2Pix. Specifically, for image editing, the paper introduces a new dataset, Image Editing in the Wild (IEIW), created by combining multiple existing datasets. However, as shown in Figure~\ref{supp fig: image editing 1}, ~\ref{supp fig: image editing 2}, ~\ref{supp fig: image editing 3}, ~\ref{supp fig: image editing 4}, ~\ref{supp fig: image editing 5}, it often produces exaggerated editing results.

\textbf{MGIE}. MLLM-Guided Image Editing (MGIE) highlights that the reliance on CLIP text encoders in Stable Diffusion limits the ability to follow precise instructions for achieving specific editing goals. To address this, MGIE replaces the CLIP text encoder with outputs from multimodal large language models (MLLMs), enabling the understanding of more expressive and detailed instructions. In contrast to MGIE, our method does not rely on a diffusion model, resulting in significantly lower complexity and avoiding the challenges of jointly optimizing models. As shown in Table~\ref{supp tab: image editing}, despite its simplicity, our approach delivers significantly better reconstruction and editing quality.

\textbf{SEED-X-Edit}. SEED-X is a unified and versatile model that can handle both comprehension and generation tasks, showcasing strong performance in real-world applications across various domains, e.g., instructed image editing. SEED-X-Edit refers to the model derived by fine-tuning SEED-X specifically for image editing on the SEED-Data-Edit dataset, a new dataset containing both manual annotated data and automatically generated image pairs. 
As shown in Table 1 and Figures~\ref{supp fig: image editing 1}, ~\ref{supp fig: image editing 2}, ~\ref{supp fig: image editing 3}, ~\ref{supp fig: image editing 4}, ~\ref{supp fig: image editing 5}, the method struggles with reconstructing unedited regions and often produces unrealistic images with high contrast. In comparison, our proposed method achieves better overall performance, despite being trained without manual-annotated data and with a simpler design.

\textbf{Inversion-based Methods}. Unlike feed-forward methods, inversion-based approaches first invert an image into a latent space, typically using latent noise or embeddings, before performing content-preserving sampling to generate the edited target. These methods require not only an editing instruction but also a source prompt, which is crucial for the inversion process. For example, Prompt-to-Prompt uses DDIM inversion and manipulates attention maps to preserve content across various edits. Similarly, Pix2Pix-Zero retains the cross-attention maps of the input image throughout the diffusion process for improved reconstruction. However, compared to our approach, these methods struggle to preserve the background. To enhance non-rigid editing, MasaCtrl modifies self-attention in diffusion models into mutual self-attention, enabling effective blending of local content and textures from input images during generation. While specialized for non-rigid edits, it falls short when applied to a variety of editing tasks. As indicated by the CLIP similarity, our method achieves better overall responsiveness to edits.

For improved reconstruction quality, optimization-based inversion methods like Null-text Inversion and PnP Inversion are proposed to invert the conditioning image into a latent embedding, achieving near-perfect reconstruction. Note the numbers in Table~\ref{supp tab: image editing} for both methods are produced with Prompt-to-Prompt. Although these methods generate higher-quality visuals, they require additional computation and time to optimize the latent embeddings. As shown in Figures~\ref{supp fig: image editing 1}, ~\ref{supp fig: image editing 2}, ~\ref{supp fig: image editing 3}, ~\ref{supp fig: image editing 4}, ~\ref{supp fig: image editing 5}, these methods are still limited by the lack of a unified, content-preserving model that consistently performs well across diverse tasks, thus limiting their scope.

\begin{figure*}[t]
    \centering
    \includegraphics[width=\linewidth]{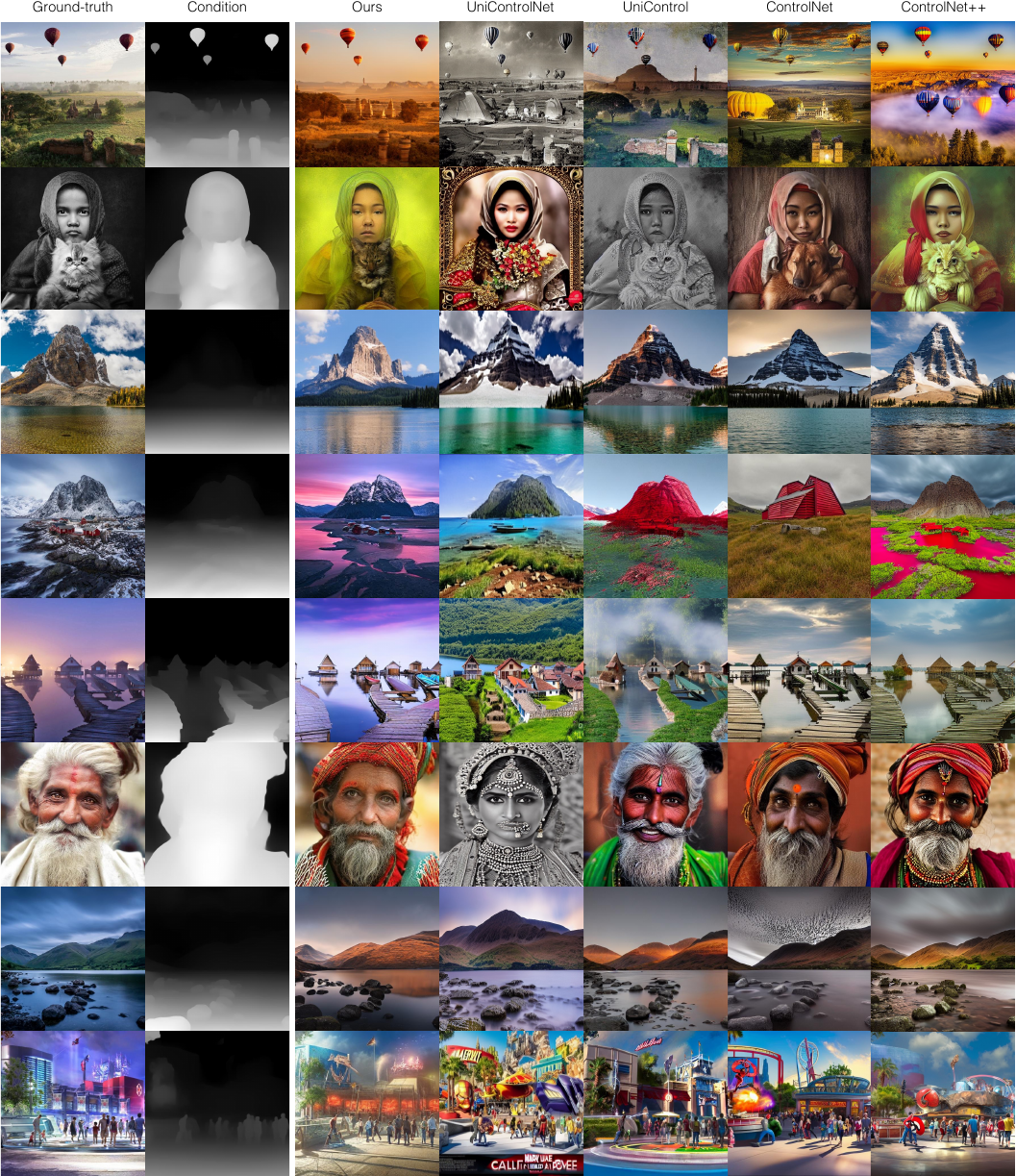}
    \caption{Comparisons complementing Figure 4. Visual comparisons to baseline methods on various depth-to-image translation. Our method, EditAR, produces photo-realistic results, preserves input details, and offers substantial sample diversity.}
    \label{supp fig: image translation depth}
\end{figure*}

\begin{figure*}[t]
    \centering
    \includegraphics[width=\linewidth]{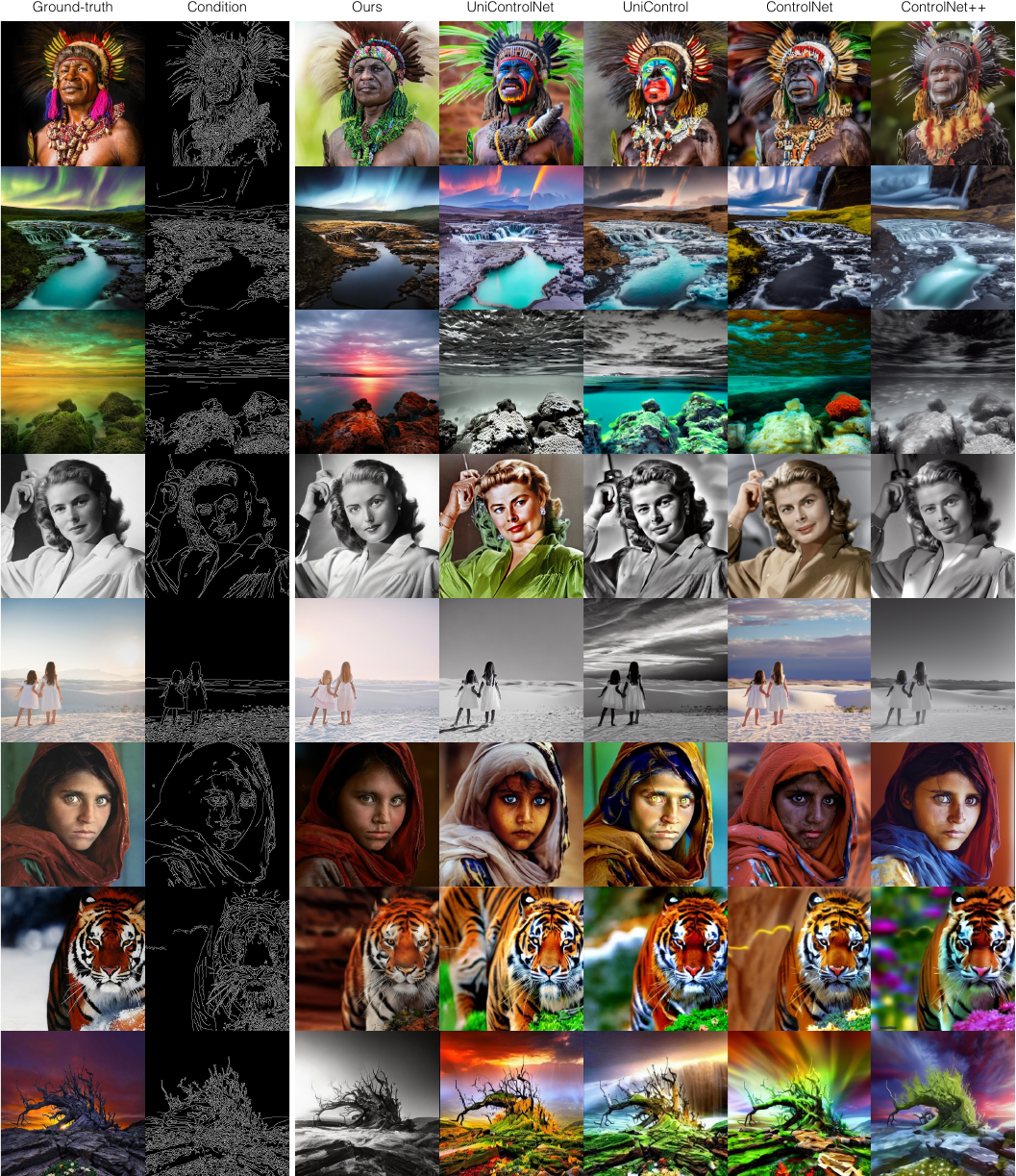}
    \caption{Comparisons complementing Figure 4. Visual comparisons to baseline methods on various edge-to-image translation. Our method, EditAR, produces photo-realistic results, preserves input details, and offers substantial sample diversity.}
    \label{supp fig: image translation canny}
\end{figure*}

\begin{figure*}[t]
    \centering
    \includegraphics[width=\linewidth]{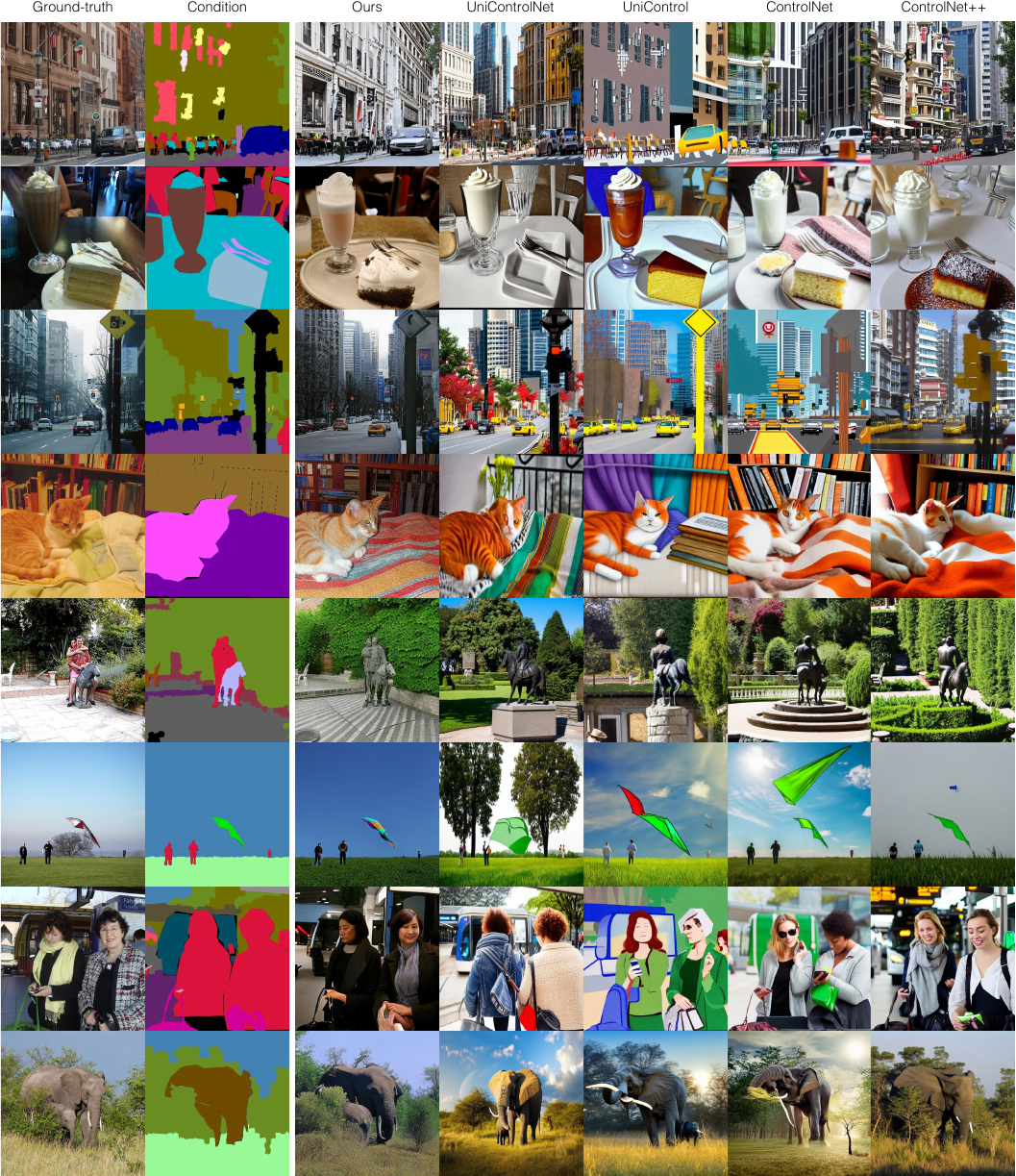}
    \caption{Comparisons complementing Figure 4. Visual comparisons to baseline methods on various segmentation-to-image translation. Our method, EditAR, produces photo-realistic results, preserves input details, and offers substantial sample diversity.}
    \label{supp fig: image translation seg}
\end{figure*}

\section{Additional Image Translation Comparison}
\label{supp section: additional image translation comparison}

In Table 2, Figure 4 and Section 4.3 in the main text, we have shown our image translation results as well as comparisons to various baselines. Here we provide more details show more visual comparisons: depth-to-image in Figure~\ref{supp fig: image translation depth}, edge-to-image in Figure~\ref{supp fig: image translation canny}, segmentation-to-image in Figure~\ref{supp fig: image translation seg}. ControlNet results are produced with ControlNet v1.1. For UniControlNet, Unicontrol and ControlNet++, we use their official released checkpoints. For each method, we produce 5,000 examples of resolution $512 \times 512$ to measure the corresponding metrics. Results show that, though learning a more challenging task, our model still synthesizes diverse images with good visual quality.

\section{Distillation}
\label{supp section: additional distillation comparison}
In Section 4.4 of the main text, we qualitatively show that adding the distillation loss improves the overall text-to-image alignment, e.g., better localizing the target editing object, on the task of image editing. For image translation, our results show that the FID scores are improved from 16.35 to 15.97 for depth-to-image, 14.43 to 13.91 for edge-to-image, 16.52 to 16.13 for segmentation-to-image.  These results further emphasize the importance of enforcing a stronger feature space similarity between the autoregressive model and foundation models, leading to models with stronger performance across tasks.

\section{Implementation Details}
\label{supp section: implementation details}



\textbf{Evaluation and Metrics}.
For image editing, the PIE-Bench dataset is used for evaluation. Specifically, PIE-Bench contains 700 images featuring ten distinct editing types: (0) random editing, (1) change object, (2) add object, (3) delete object, (4) change object content, (5) change object pose, (6) change object color, (7) change object material, (8) change background, and (9) change image style. 
Within each scene, images are evenly distributed among four categories: animal, human, indoor environment, and outdoor environment.
Our method as well as all other feed-forward methods uses the source image and editing instructions to predict the target edit. The inversion-based approaches use the source image, the source prompt, and the target image prompt. Structure Distance ($\times 10^3$) leverages self-similarity of DINO-ViT features and computes cosine similarity between image features as structure distance. PSNR, LPIPS ($\times 10^3$), MSE ($\times 10^4$), and SSIM ($\times 10^2$) are reported to compare the background preservation using the manual-annotated masks. The CLIP score ($\times 10^2$) evaluates text-image similarity between the edited images and corresponding target editing text prompts. Both the whole image and regions in the editing mask (black out everything outside the mask) are calculated, and referred to as Whole Image Clip and Edit Region Clip, respectively. All metrics are computed at the resolution of $512 \times 512$.

For the evaluation of image translation, we follow ControlNet++ and use the corresponding validation splits for COCOStuff and MultiGen-20M, which contain $5,000$ examples per task. Regarding metrics, we follow the common practice in the field: mIOU ($\times 10^2$) is used for semantic segmentation conditions, RMSE for depth map conditions, and SSIM ($\times 10^2$) for canny edge conditions. FID scores are computed with $5,000$ images at the resolution of $512 \times 512$.

\textbf{Training and Inference}. To overcome varying imbalances between tasks, datasets must be mixed thoughtfully. We mix datasets by sampling $15\%$ for each image translation task, $25\%$ for PIPE dataset, and $30\%$ for SEED-Data-Edit-Unsplash. The training hyperparameters mostly follow LlamaGen. All images are resized to a resolution of $512 \times 512$ for both training and inference. The VQ-Autoencoder has a downsampling ratio of 16, so that each image is represented by 1024 tokens. Its dictionary size is 16384 and embedding dimensionality is 8. The text encoder utilizes Flan-T5-XL, producing a sequence of 120 embeddings. We use the pre-trained text-to-image autoregressive model LlamaGen GPT-XL, which has 36 layers and an embedding dimension of 1280. The model is optimized using AdamW with a constant learning rate of $10^{-4}$, $\beta_1 = 0.9$, $\beta_2 = 0.95$, and weight decay of $0.05$. The model is trained with a batch size of 64 for $40,000$ iterations on 8 A100 GPUs. We use $\lambda_{distill}=0.5$ and $\eta = 3.0$ for inference.

\section{Discussion}
\label{supp section: discussion}

EditAR is a versatile autoregressive model that unifies multiple conditional image generation tasks within a single framework. Using only text prompts, the model seamlessly adapts to various image inputs and tasks. Our comprehensive evaluation demonstrates EditAR's exceptional performance in both image editing and diverse image translation tasks. This work represents a significant milestone as the first demonstration that a single autoregressive model using next-token prediction can effectively handle various conditional generation tasks on large-scale benchmarks. By successfully tackling multiple conditional image generation challenges, EditAR opens new possibilities for unified conditional generation approaches, building upon recent advances in text-to-image autoregressive modeling.

\textbf{Limitations.} EditAR builds upon autoregressive text-to-image models, allowing it to naturally benefit from advances in base model quality. Besides, the current implementation is restricted to single-image conditional inputs, though the framework could theoretically handle multiple conditions. Additionally, the model struggles with non-rigid or 3D editing tasks due to the insufficient training data. Addressing these challenges through expanded datasets and architectural enhancements represents an important direction for future research.




\end{document}